\documentclass[10pt, oneside]{article}    

\usepackage{geometry}
\usepackage{graphicx}
\usepackage{color}
\usepackage{xcolor}
\usepackage{amssymb}
\usepackage{hyperref}
\usepackage{comment}
\usepackage{enumitem}

\definecolor{darkblue}{rgb}{0.0,0.0,0.3}
\hypersetup{colorlinks,breaklinks,linkcolor=darkblue,urlcolor=darkblue,anchorcolor=darkblue,citecolor=darkblue}

\usepackage{sectsty}
\sectionfont{\fontsize{12}{14}\selectfont}

\usepackage{longtable}

\usepackage{booktabs}

\title{The SpaceNet Multi-Temporal Urban Development Challenge}
\date{\vspace{-4ex}}

 \author{Adam {Van Etten}\thanks{In-Q-Tel CosmiQ Works, [avanetten, dhogan]@iqt.org}  \thanks{Thanks to all of the SpaceNet Partners: CosmiQ Works, Maxar Technologies,  Amazon Web Services, Capella Space, Topcoder, IEEE GRSS, the National Geospatial-Intelligence Agency, Planet.
Special thanks to Nick Weir for project design and Jesus Martinez-Manso for dataset curation.} \and Daniel Hogan\footnotemark[1]
}

\begin{document}

\maketitle

\begin{abstract}

Building footprints provide a useful proxy for many humanitarian applications.  For example, building footprints are useful for high fidelity population estimates, and quantifying population statistics is fundamental to $\sim1/4$ of the United Nations Sustainable Development Goals Indicators.  In this paper we (the SpaceNet Partners) discuss efforts to develop techniques for precise building footprint localization, tracking, and change detection via the SpaceNet Multi-Temporal Urban Development Challenge (also known as SpaceNet 7).  
 In this NeurIPS 2020 competition, participants were asked identify and track buildings in satellite imagery time series collected over rapidly urbanizing areas. The competition centered around a brand new open source dataset of Planet Labs satellite imagery mosaics at 4m resolution, which includes 24 images (one per month) covering $\approx100$ unique geographies. 
 Tracking individual buildings at this resolution is quite challenging, yet the winning participants demonstrated impressive performance with the newly developed SpaceNet Change and Object Tracking (SCOT) metric.  This paper details the top-5 winning approaches, as well as analysis of results that yielded a handful of interesting anecdotes such as decreasing performance with latitude.

\end{abstract}


\section{Background}

Time series analysis of satellite imagery poses an interesting computer vision challenge with numerous human development applications. The SpaceNet 7 Multi-Temporal Urban Development Challenge aims to advance this field through a data science competition aimed specifically at improving these methods. Beyond its relevance for disaster response, disease preparedness, and environmental monitoring, this task poses technical challenges currently unaddressed by existing methods. SpaceNet is a nonprofit LLC dedicated to accelerating open source, artificial intelligence applied research for geospatial applications, specifically foundational mapping ({\it i.e.} building footprint \& road network detection).   

From 2016 - March 2021, SpaceNet was run by co-founder and managing partner CosmiQ Works, in collaboration with co-founder and co-chair Maxar Technologies and partners including Amazon Web Services (AWS), Capella Space, Topcoder, IEEE GRSS, the National Geospatial-Intelligence Agency and Planet.
The SpaceNet Multi-Temporal Urban Development Challenge represents the seventh iteration of the SpaceNet Challenge series, in which each challenge addresses a previously ill-understood aspect of geospatial data analysis. This was the first SpaceNet Challenge to involve a time series element.
In this section we detail the impacts, both technical and social, of the SpaceNet 7 Challenge.

In this competition we challenged participants to identify new building construction in satellite imagery, which could enable development policies and aid efforts by improving population estimation. High-resolution population estimates help identify communities at risk for natural and human-derived disasters. Population estimates are also essential for assessing burden on infrastructure, from roads \cite{road_crashes} to medical facilities \cite{hospital_catchment} and beyond. Organizations like the World Bank and the World Health Organization use these estimates when evaluating infrastructure loans, grants, and other aid programs \cite{WB_annual_report}. However, population estimates are often inaccurate, out-of-date, or non-existent in many parts of the world. In 2015, the World Bank estimated that 110 countries globally lack effective systems for Civil Registration and Vital Statistics (CRVS), \textit{i.e.} birth, death, marriage, and divorce registration \cite{mills_2015}. CRVS are also fundamental to assessing progress in 67 of the 231 UN Sustainable Development Goals indicators \cite{WB_CRVS_SGD}. Inaccurate population estimates can result in poor distribution of government spending and aid distribution, overcrowded hospitals, and inaccurate risk assessments for natural disasters \cite{Guha-Sapir}.  

Importantly, the computer vision lessons learned from this competition could apply to other data types. Several unusual features of satellite imagery (\textit{e.g.} small object size, high object density, different color band wavelengths and counts, limited texture information, drastic changes in shadows, and repeating patterns) are relevant to other tasks and data. For example, pathology slide images or other microscopy data present all of the same challenges \cite{vwtas}. Lessons learned in the SpaceNet Multi-Temporal Urban Development Challenge may therefore have broad-reaching relevance to the computer vision community.

\section{Novelty}\label{sec:novelty}

\begin{figure}
    \centering
    \includegraphics[width=0.85\linewidth]{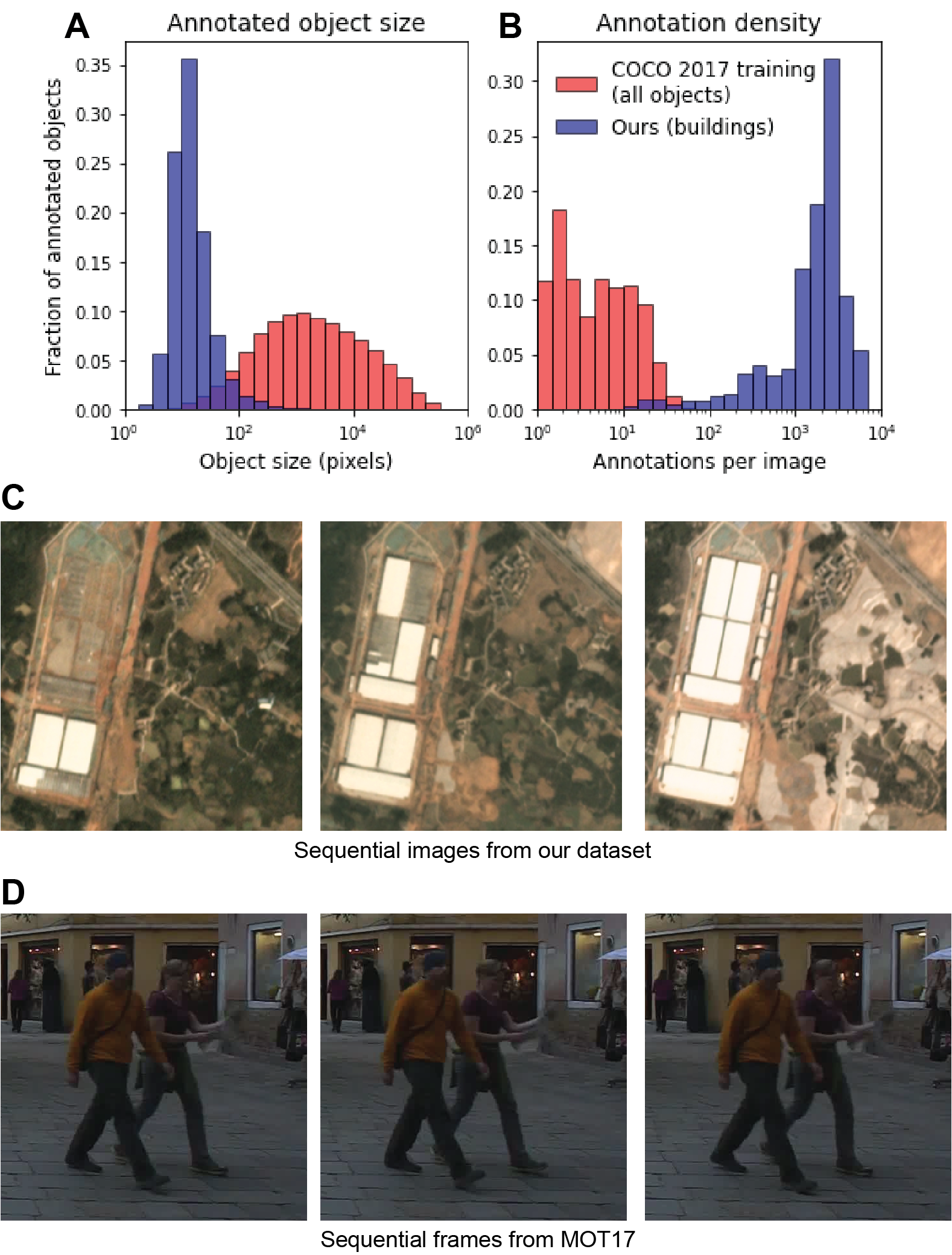}
    \caption{\textbf{A comparison between our dataset and related datasets.} \textbf{A.} Annotated objects are very small in this dataset. Plot represents normalized histograms of object size in pixels. Blue is our dataset, red represents all annotations in the COCO 2017 training dataset \cite{coco}. \textbf{B.} The density of annotations is very high in our dataset. In each $1024 \times 1024$ image, our preliminary dataset has between 10 and over 20,000 objects (mean: 4,600). By contrast, the COCO 2017 training dataset has at most 50 objects per image. \textbf{C.} Three sequential time points from one geography in our dataset, spanning 3 months of development. Compare to \textbf{D.}, which displays three sequential frames in the MOT17 video dataset \cite{MOT17}.}
    \label{fig:comparison}
\end{figure}

Past data science competitions have not studied deep time series of satellite imagery. The closest comparison is the xView2 challenge \cite{xbd}, which examined building damage in satellite image pairs acquired before and after natural disasters; however, this task fails to address the complexities and opportunities posed by analysis of deep time series data, such as seasonal foliage and lighting changes. Another related dataset/challenge is Functional Map of the World \cite{fmow} (which is hosted by SpaceNet).  This dataset contains some temporal information, though time series are irregular (a plurality of locations have only a single observation), and the task is static satellite scene classification rather than dynamic object tracking as in SpaceNet 7.  
Other competitions have explored time series data in the form of natural scene video, \textit{e.g.} object detection \cite{MOT17} and segmentation \cite{Caelles_arXiv_2019} tasks. There are several meaningful dissimilarities between these challenges and the competition described here. For example, frame-to-frame variation is very small in video datasets (see Figure \ref{fig:comparison}D). By contrast, the appearance of satellite images can change dramatically from month to month due to differences in weather, illumination, and seasonal effects on the ground, as shown in Figure \ref{fig:comparison}C. Other time series competitions have used non-imagery data spaced regularly over longer time intervals \cite{kaggle_google_traffic}, but none focused on computer vision tasks.

The challenge built around the VOT dataset \cite{VOT_TPAMI} saw impressive results for video object tracking ({\it e.g.} \cite{Wang_2019_CVPR}), yet this dataset differs greatly from satellite imagery, with high frame rates and a single object per frame.  Other datasets such as MOT17 \cite{MOT17} have multiple targets of interest, but still have relatively few ($< 20$) objects per frame. The Stanford Drone Dataset \cite{Robicquet2016LearningSE} appears similar at first glance, but has several fundamental differences that result in very different applications. That dataset contains overhead videos taken at multiple hertz  from a low elevation, and typically have $\approx20$ moving objects (cars, people, buses, bicyclists, etc.) per frame. Because of the high frame rate of these datasets, frame-to-frame variation is minimal (see the MOT17 example in Figure \ref{fig:comparison}D). Furthermore, objects are larger and less abundant in these datasets than buildings are in satellite imagery. As a result, video competitions and models derived therein provide limited insight in how to manage imagery time series with substantial image-to-image variation. Our competition and data address this gap (see Section \ref{sec:novelty} and Section \ref{sec:data}).

The size and density of target objects are very different in this competition than past computer vision challenges. When comparing the size of annotated instances in the COCO dataset \cite{coco}, there's a clear difference in object size distributions (see Figure \ref{fig:comparison}A). These smaller objects intrinsically provide less information as they comprise fewer pixels, making their identification a more difficult task. Finally, the number of instances per image is markedly different in satellite imagery from the average natural scene dataset (see Section \ref{sec:data} and Figure \ref{fig:comparison}B).  Other data science competitions have explored datasets with similar object size and density, particularly in the microscopy domain \cite{recursion, BAH}; however, those competitions did not address time series applications.

\section{Data}\label{sec:data}

In this section we briefly detail the dataset used in SpaceNet 7; for a detailed description of the Multi-temporal Urban Development SpaceNet (MUDS) dataset and baseline algorithm, see \cite{sn7_muds}.  The SpaceNet 7 Challenge used a brand-new, open source dataset of medium-resolution ($\approx 4$ m) satellite imagery collected by Planet Labs' Dove Satellites between 2017 and 2020. The dataset is  open sourced under the CC-BY-4.0 ShareAlike International license.  
As part of AWS's Open Data Program\footnote{https://registry.opendata.aws/spacenet/}, SpaceNet data is entirely free to download.

The imagery comprises 24 consecutive monthly mosaic images (a mosaic is a combination of images stitched together, often made to minimize cloud cover) of 101 locations  over 6 continents, totaling $\approx 40,000 \, \rm{km}^{2}$ of satellite imagery. The dataset's total imaged area compares favorably to past SpaceNet challenge datasets, which covered between $120 \, \rm{km}^{2}$ and $3,000 \, \rm{km}^{2}$ \cite{spacenet_orig, etten2020road, MVOI}. 

Each image in the dataset is accompanied by two sets of manually created annotations. The first set are GeoJSON-formatted, geo-registered building footprint polygons defining the precise outline of each building in the image.  Each building is assigned a unique identifier that persists across the time series. The second annotations, provided in the same format, are ``unusable data masks'' (UDMs) denoting areas of images obscured by clouds. Each $1024 \times 1024$ image has between 10 and $\approx20,000$ building annotations, with a mean of $\approx4,600$ (the earliest timepoints in some geographies have very few buildings completed). This represents much higher label density than natural scene datasets like COCO \cite{coco} (Figure \ref{fig:comparison}B), or even overhead drone video datasets \cite{stanford_drone}. 

\begin{figure}
    \centering
    \vspace{-5pt}
    \includegraphics[width=0.99\linewidth]{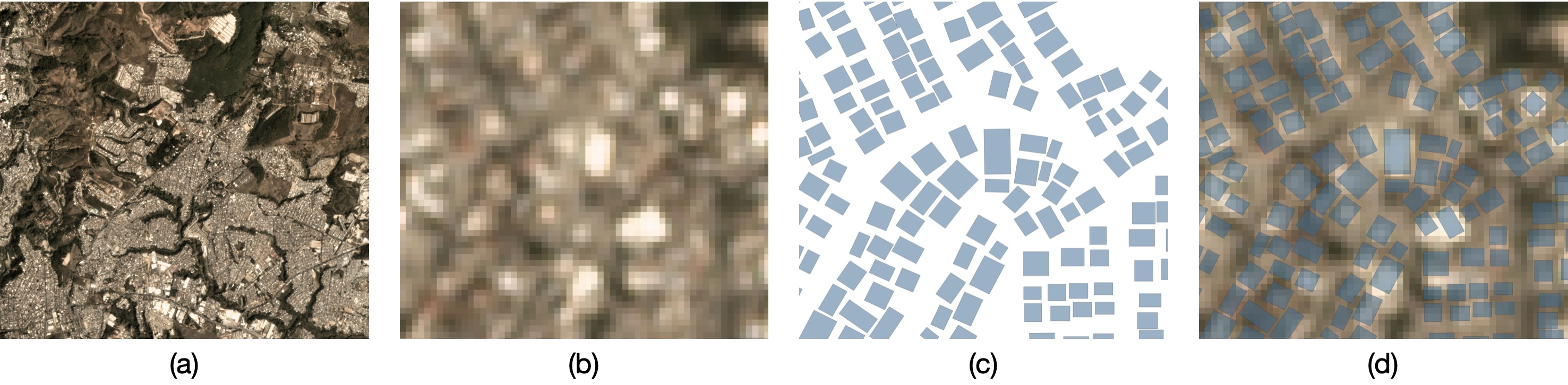}
    \vspace{-5pt}
    \caption{Zoom-in of one particularly dense SpaceNet 7 region illustrating the very high fidelity of labels. (a) Full image. (b) Zoomed cutout. (c) Footprint polygon labels. (d) Footprints overlaid on imagery.}
    \label{fig:omni}
    \vspace{-8pt}
\end{figure}

The labeling process for SpaceNet 7 was an exhaustive 7-month effort that utilized both the native Planet 4m resolution imagery, as well as higher-resolution imagery in particularly difficult scenes.  By leveraging complementary data sources, the labelers were able to create what we have dubbed ``omniscient'' labels that appear to be far higher quality than what the imagery merits.  Figure \ref{fig:omni}  illustrates that in some dense scenes, label precision exceeds what the human eye could easily distinguish in 4m resolution imagery. 

The final dataset includes $\approx$11M annotations, representing $\sim500,000$ unique buildings. For the challenge, we released 60 of the 101 AOIs (area of interest, i.e., location) for training; this portion included both imagery and labels. Imagery (not labels) for 20 of the AOIs were released as the ``test\_public''.  The remaining 21 AOIs were withheld as the ``test\_private'' set.  Taken together, the test set includes 4.4 million annotated buildings.

\section{Metric}\label{sec:metrics}

For this competition we defined successful building footprint identifications as proposals which overlap ground truth ($GT$) annotations with an Intersection-over-Union ($IoU$) score above a threshold of 0.25. The $IoU$ threshold here is lower than the $IoU \geq 0.5$ threshold of previous SpaceNet challenges \cite{MVOI, spacenet_orig, etten2020road} due to the increased difficulty of building footprint detection at reduced resolution and very small pixel areas. 

To evaluate model performance on a time series of identifier-tagged footprints, we introduce a new evaluation metric: the SpaceNet Change and Object Tracking (SCOT) metric.  See \cite{sn7_muds} for further details.  
In brief, the SCOT metric combines two terms: a tracking term and a change detection term. The tracking term evaluates how often a proposal correctly tracks the same buildings from month to month with consistent identifier numbers. In other words, it measures the model's ability to characterize what stays the same as time goes by. The change detection term evaluates how often a proposal correctly picks up on the construction of new buildings. In other words, it measures the model's ability to characterize what changes as time goes by.  
The combined tracking and change terms of SCOT therefore provide a good measure of the dynamism of each scene.

\section{Challenge Structure}\label{sec:chall}

\begin{figure*}
    \centering
    \includegraphics[width=0.99\linewidth]{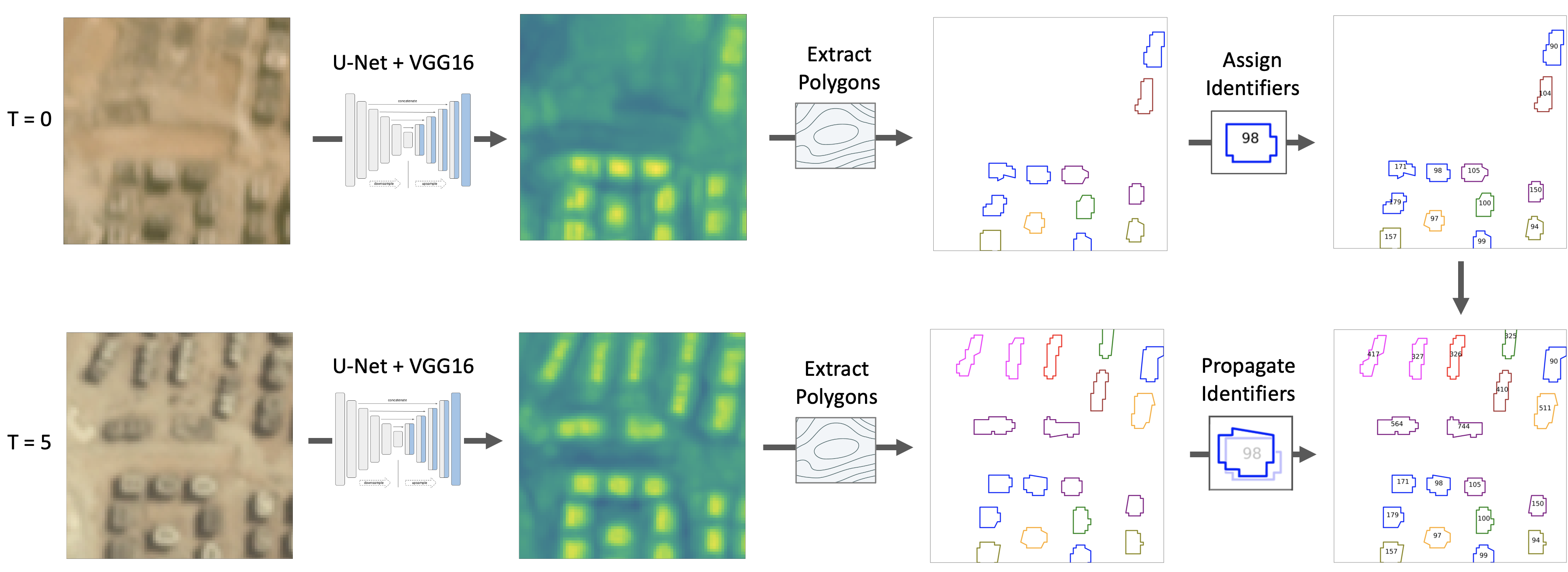}
    \vspace{-3pt}
    \caption{Baseline algorithm for building footprint extraction and identifier tracking showing evolution from T = 0 (top row) to T = 5 (bottom row).  Imagery (first column) feeds into the segmentation model, yielding a building mask (second column).  This mask is refined into building footprints (third column), and unique identifiers are allocated (right column).}
    \label{fig:baseline}
\end{figure*}

The competition focused on a singular task: tracking building footprints to monitor construction and demolition in satellite imagery time series.  Beyond the training data, a baseline model\footnote{https://github.com/CosmiQ/CosmiQ\_SN7\_Baseline} was provided to challenge participants
to demonstrate the feasibility of the challenge task.
This challenge baseline used a state-of-the-art building detection algorithm adapted from one of the prize winners in the SpaceNet 4 Building Footprint Extraction Challenge \cite{MVOI}. 
Binary building prediction masks are converted to instance segmentations of building footprints. Next, footprints at the same location over the time series are be assigned the same unique identifier, see Figure \ref{fig:baseline}.

The effects and challenges associated with population estimates are myriad and very location-dependent, and it is therefore critical to involve scientists in areas of study who rarely have access to these data. To this end, the SpaceNet partners worked hard to lower the barrier of entry for competing: firstly, all data for this challenge is free to download.
Secondly, the SpaceNet partners provided \$25,000 in AWS compute credits to participants to enable data scientists without extensive personal compute resources to compete. To enhance the value of these two enabling resources and to further increase engagement with affected communities, 
we provided extensive tutorial materials on The DownLinQ\footnote{https://medium.com/the-downlinq} detailing how to download data, prepare data, run the baseline model, utilize AWS credits, and score output predictions.   We used an internationally known competition hosting platform to ensure accessibility of the challenge worldwide (Topcoder).

The challenge
ran from September 8, 2020 - October 28, 2020.  
An initial leaderboard for the 311 registrants was based upon predictions submitted for the ``test\_public'' set.  The top 10 entries on this leaderboard at challenge close were invited to submit their code in a Docker container.  The top 10 models were subsequently retrained (to ensure code was working as advertised), and then internally tested on the ``test\_private'' set of 21 new geographies.  This step of retraining the models and testing on completely unseen data minimizes the chances of cheating, and ensures that models are not hypertuned for the known test set.  The scores on the withheld ``test\_private'' set determine the final placings, with the winners announced on December 2, 2020. 
A total of $\$$50,000 USD  was awarded to the winners 
(1st=\$20,000 USD, 2nd=\$10,000 USD, 3rd=\$7,500 USD, 4th=\$5,000 USD, 5th=\$2,500 USD, Top Graduate=\$2,500 USD, Top Undergraduate=\$2,500 USD).  
The top-5 winning algorithms are open-sourced under a permissive license\footnote{https://github.com/SpaceNetChallenge/SpaceNet7\_Multi-Temporal\_Solutions}.


\section{Overall Results}

SpaceNet 7 winning submissions applied varied techniques to solving the challenge task, with the most creativity reserved to post-processing techniques (particularly the winning implementation, see Section \ref{sec:winning}) .  
Notably, post-processing approaches did not simply rely upon the tried-and-true fallback of adding yet another model to an ensemble. In fact, the winning model did not use an ensemble of neural network architectures at all, and managed an impressive score with only a single, rapid model. 
Table \ref{tab:results} details the top-5 prize winning competitors of the 300+ participants in SpaceNet 7.

\begin{figure}
    \centering
    \includegraphics[width=0.7\linewidth]{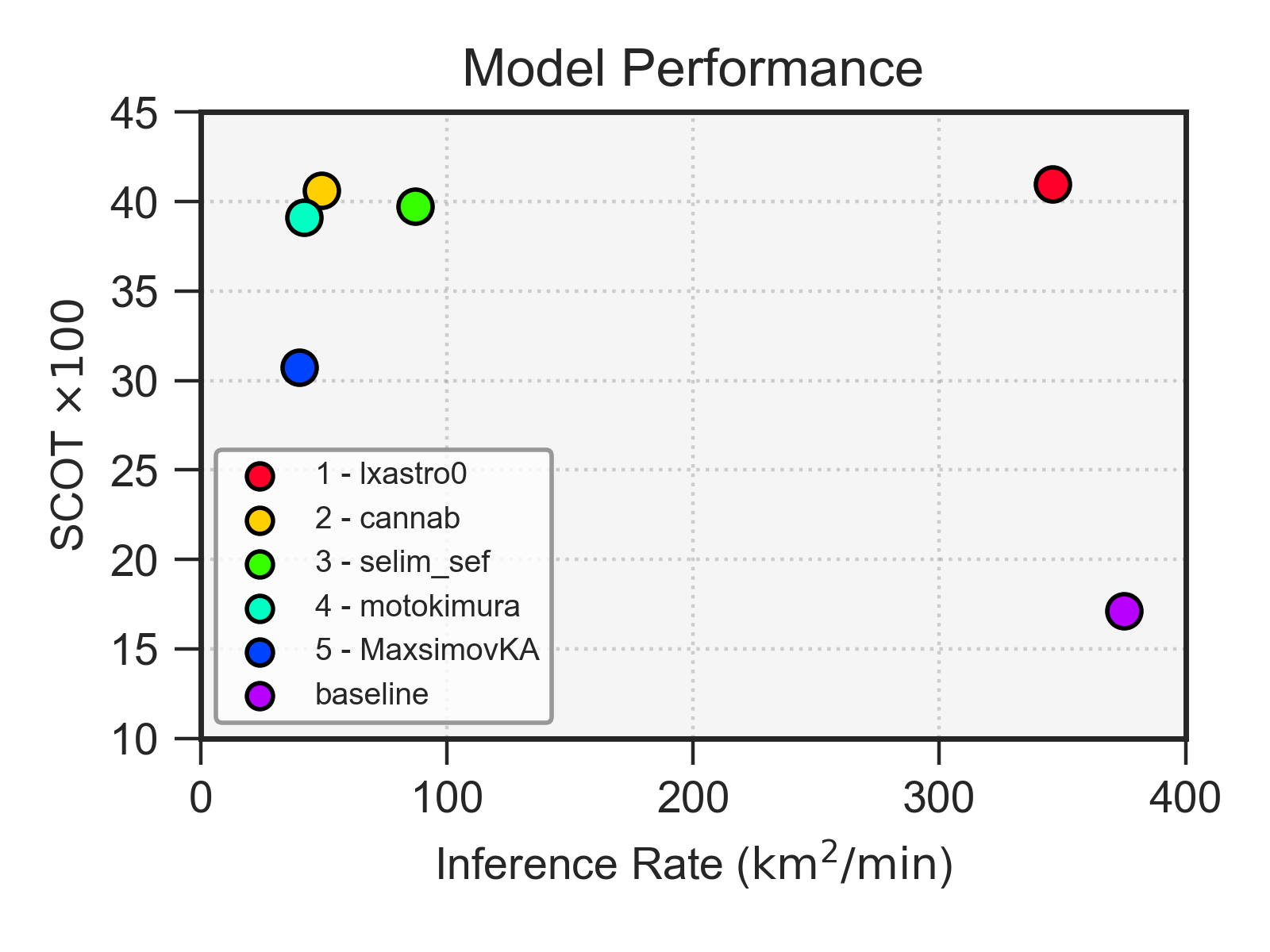}
    \vspace{-8pt}
    \caption{Performance vs speed for the winning algorithms.  Up and to the right is best; the1st place algorithm is many times faster than the runner-up submissions.}
    \label{fig:rate}
    \vspace{-8pt}
\end{figure}

We see from Table \ref{tab:results} that ensembles of models are not a panacea, and in fact post-processing techniques have a far greater impact on performance than the individual architecture chosen.  The winning algorithm is a clear leader when it comes to the combination of performance and speed, as illustrated in Figure \ref{fig:rate}.  

\begin{table*}
  \caption{SpaceNet 7 Results}
  \label{tab:test_regs}
  \footnotesize
  \centering
   \begin{tabular}{llllllll}
    \hline
    {\bf Competitor} & {\bf Final} & {\bf Total} & {\bf Architectures} & {\bf \# }  & {\bf Training}  & {\bf Speed}\\
    			     & {\bf Place} & {\bf Score} & 		    	         & {\bf Models}	& {\bf Time (H)} & {\bf  (km$^2$/min)}\\
    \hline
   lxastro0		     & 1		& 41.00	    & 1 $\times$ HRNet &1			  	& 36			 & 346 \\
   cannab		     & 2		& 40.63	    & 6 $\times$ EfficienNet + UNet (siamese)  &  6	& 23	 & 49 \\
   selim\_sef	     &3		& 39.75	    & 4 $\times$ EfficienNet + UNet &    4 	 & 46			  & 87  \\
   motokimura	     &4		& 39.11	    & 10 $\times$ EfficienNet-b6 + UNet 	& 10	& 31		 & 42 \\
   MaxsimovKA	     & 5		& 30.74	    & 1 $\times$ SENet154 +UNet (siamese) & 1 & 15	 & 40 \\
   baseline		     & N/A		& 17.11	    & 1 $\times$  VGG16 + UNet	 &1 		& 10			 & 375 \\
\hline
  \end{tabular}
  \label{tab:results}
  \vspace{-5pt}
\end{table*}

\section{Segmentation Models} 

As noted above, post-processing techniques are really where the winning submissions differentiated themselves (and will be covered in depth in Section \ref{sec:winning}), but there are a few trends in the initial deep learning segmentation approach worth noting.  

\begin{enumerate}
	\item {\bf Upsampling Improved Performance} The moderate resolution of imagery poses a significant challenge when extracting small footprints, so multiple competitors upsampled the imagery $3 - 4\times$ and noted improved performance.
	\item {\bf 3-channel Training Mask} The small pixel sizes of many buildings results in very dense clustering in some locations, complicating the process of footprint extraction.  Accordingly, multiple competitors found utility in 3-channel ``footprint, boundary, contact'' (fbc\footnote{https://solaris.readthedocs.io/en/latest/tutorials/notebooks/api\_masks\_tutorial.html}) segmentation masks for training their deep learning models.
	\item {\bf Ensembles Remain the Norm} While the winning algorithm eschewed multi-model ensembles (to great speed benefits), the remainder of the top-4 competitors used an ensemble of segmentation models which were then averaged to form a final mask.  
\end{enumerate}

\section{Winning Approach}\label{sec:winning}

While there were interesting techniques adopted by all the winning algorithms, the vastly superior speed of the winning algorithm compared to the runners-up merits a closer look.
The winning team of lxastro0 (consisting of four Baidu engineers) improved upon the baseline approach in three key ways.  
\begin{enumerate}
	\item They swapped out the VGG16 \cite{vgg16} + U-Net \cite{u-net} architecture of the baseline with the more advanced HRNet \cite{hrnet}, which maintains high-resolution representations through the whole network.  Given the small size of the SpaceNet 7 buildings, mitigating the downsampling present in most architectures is highly desirable. 
	\item The small size of objects of interest is further mitigated by upsampling the imagery 3$\times$ prior to ingestion into HRNet.  The team experimented with both 2$\times$ and 3$\times$ upsampling, and found that 3$\times$ upsampling proved superior.
	\item Finally, and most crucially, the team adopted an elaborate post-processing scheme they term "temporal collapse" which we detail in Section \ref{sec:collapse}.
\end{enumerate}

\subsection{Temporal Collapse}\label{sec:collapse}

In order to improve post-processing for SpaceNet 7, the winning team assumed: 
\begin{enumerate}
	\item Buildings will not change after the first observation.
	\item In the 3$\times$ scale, there is at least a one-pixel gap between buildings.  
	\item There are three scenarios for all building candidates:
		\begin{enumerate}
			\item Always exists in all frames
			\item Never exists in any frame
			\item Appears at some frame k and persists thereafter
		\end{enumerate}
\end{enumerate}

\begin{figure}
    \centering
    \begin{tabular}{cc}
        \includegraphics[width=0.58\linewidth]{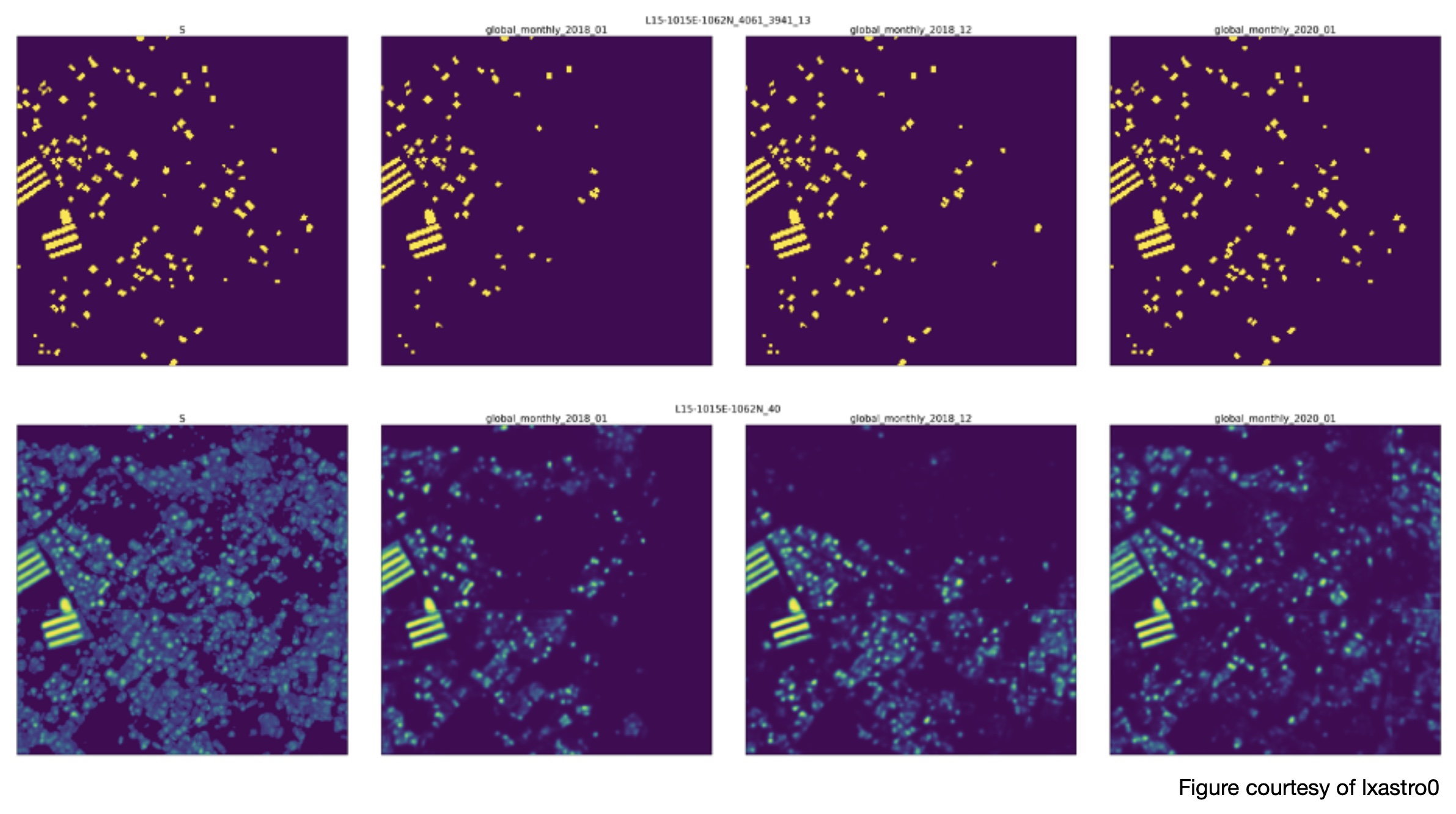} &
 	\includegraphics[width=0.39\linewidth]{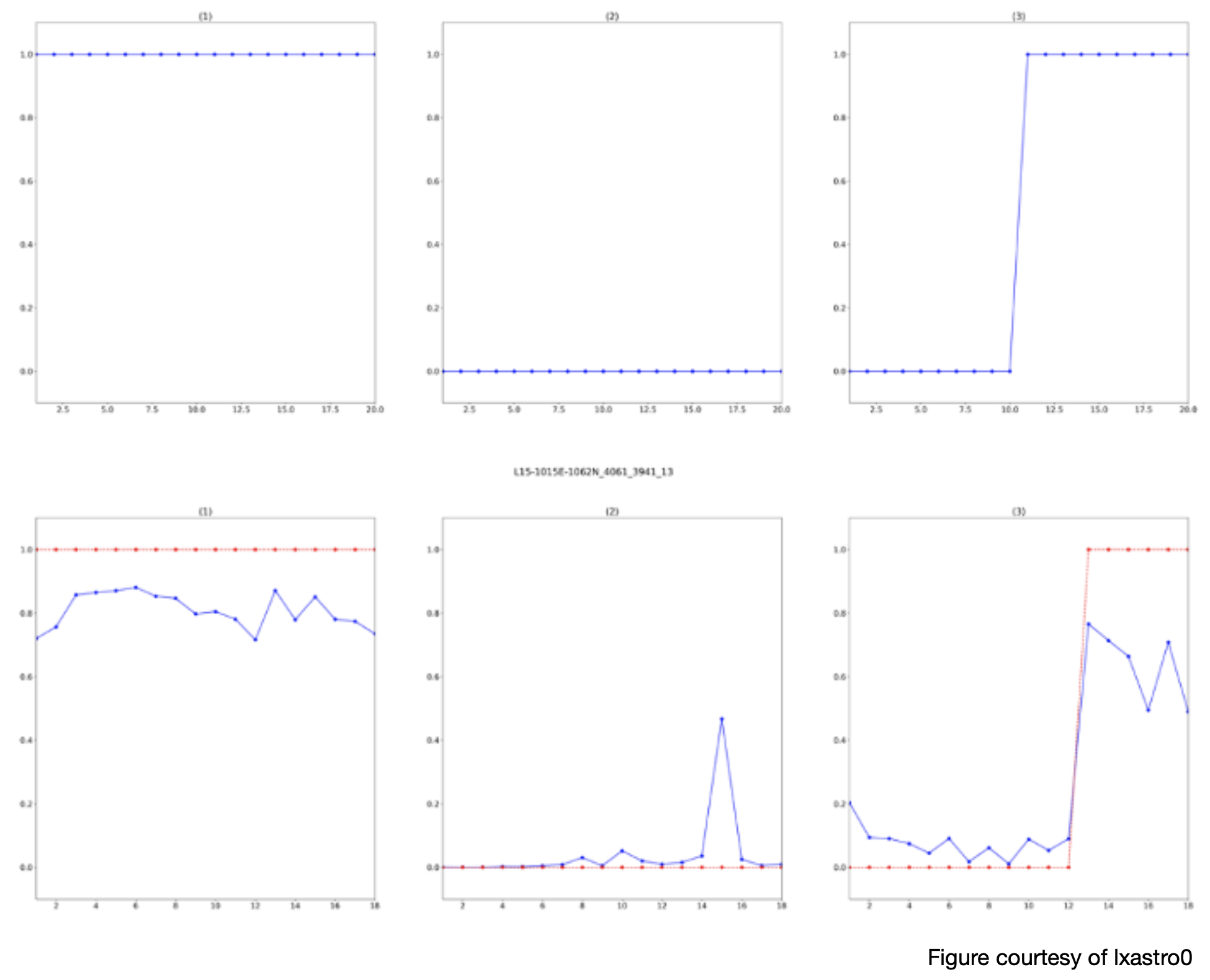} \\
    (a)  & (b)  \\ [-0pt] 
    \end{tabular}
    \vspace{-5pt}
    \caption{ {\bf (a)} Visualization of temporal collapse for ground truth (top row) and predictions (bottom row).  The left frame is the compressed probability map.
    	{\bf (b)} Method for determining the temporal origin of an individual building.  Top row: The three possible scenarios of assumption (c).  Bottom row: The aggregated predicted probability for the building footprint at each time step (blue) is used to map to the final estimated origin (red).}
    \label{fig:collapse}
    \vspace{-5pt}
\end{figure}

The data cube for each AOI  can be treated as a video with a small ($\sim24$) number of frames.  Since assumption (1) states that building boundaries are static over time, lxastro0 compresses the temporal dimension and predicts the spatial location of each building only once, as illustrated in Figure \ref{fig:collapse}a.

Building footprint boundaries are extracted from the collapsed mask using the watershed algorithm and an adaptive threshold, and taking into account assumption (2).  This spatial collapse ensures that predicted building footprint boundaries remain the same throughout the time series.  
With the spatial location of each building now determined, the temporal origin must be computed. At each frame, and for each building, the winning team averaged the predicted probability values at each pixel inside the pre-determined building boundary. This mapping is then used to determine at which frame the building originated, as illustrated in Figure \ref{fig:collapse}b.

\begin{figure}[h]
    \centering
    \begin{tabular}{cc}
    \includegraphics[width=0.5\linewidth]{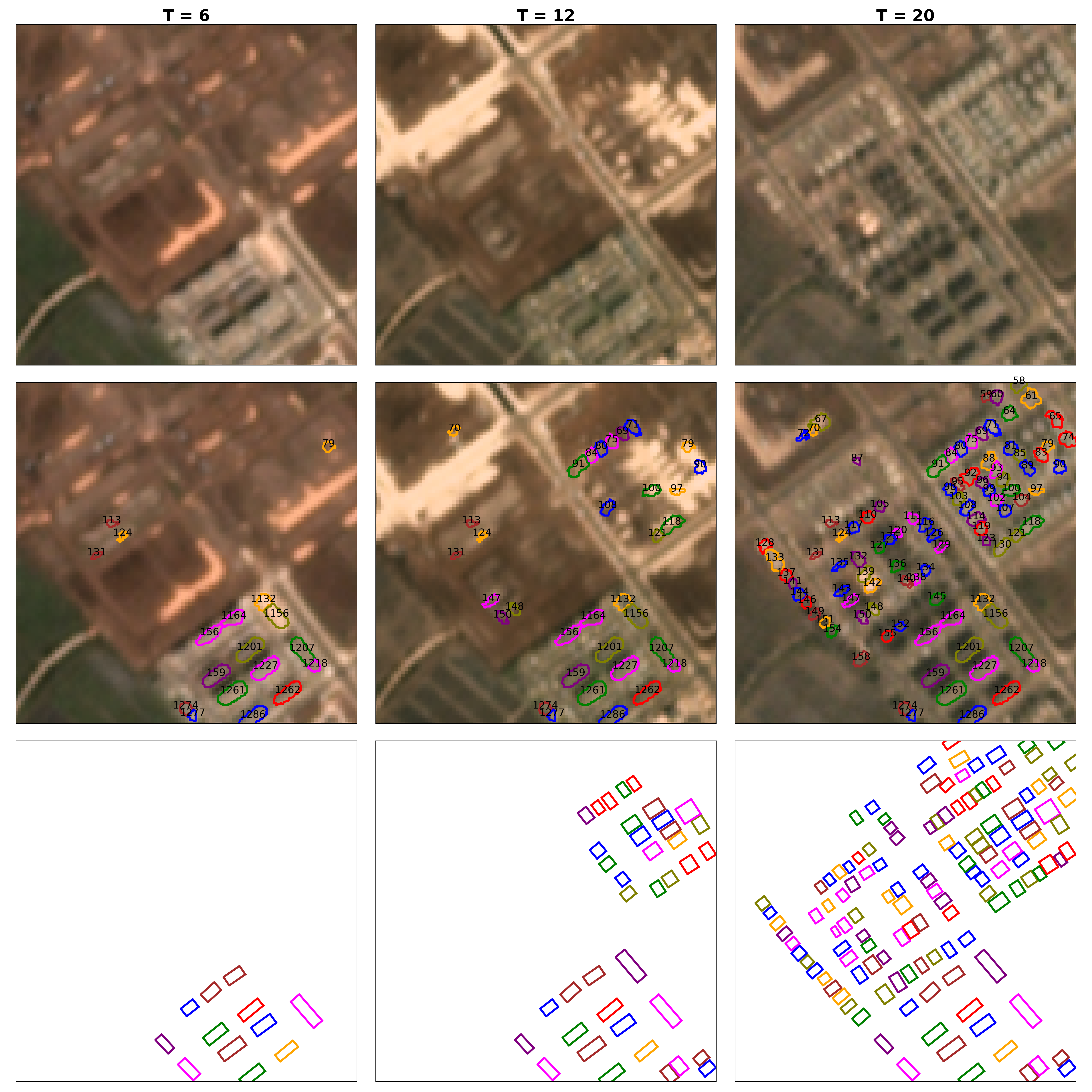} &   
    \includegraphics[width=0.5\linewidth]{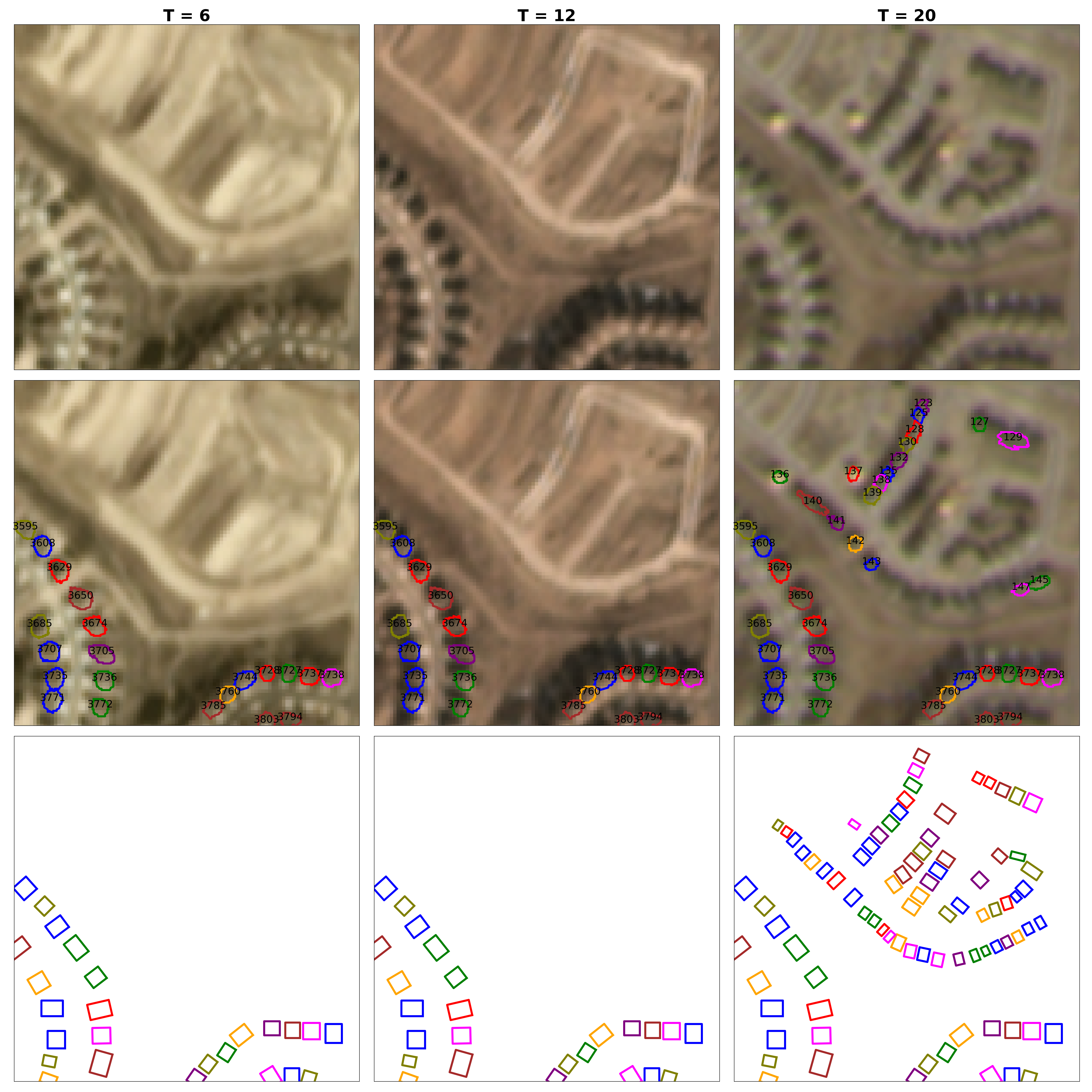} \\ 
    (a) & (b) \\ [-0pt] 
    \end{tabular}
    \vspace{-5pt}
    \caption{ {\bf Imagery, predictions and ground truth.} Input imagery (top row),  predictions (middle row), and ground truth (bottom row) of the winning model for sample test regions.  The left column denotes month 6 (October 2018), with the middle column 6 months later and the right column another 8 months later. 
    {\bf (a)} AOI 1, latitude = 20$^\circ$, change score = 0.30. 
    {\bf (b)} AOI 2, latitude = 40$^\circ$, change score = 0.09. 
    }
    \label{fig:bloob}
    \vspace{-5pt}
\end{figure}

The techniques adopted by lxastro0 yield marked improvements over the baseline model in all metrics, but most importantly in the change detection term of the SpaceNet Change and Object Tracking (SCOT) metric.  See Table \ref{tab:comp} for quantitative improvements.  
Figure \ref{fig:bloob}a illustrates predictions in a difficult region, demonstrating that while the model is imperfect, it does do a respectable job given the density of buildings and moderate resolution. We discuss Figure \ref{fig:bloob}b in Section \ref{sec:corr}.

\begin{table}[t]
  \caption{baseline model vs lxastro0}
 \vspace{3pt}
  \label{tab:comp}
  \centering
   \begin{tabular}{lll}
    \hline
    {\bf Metric} & {\bf baseline} & {\bf lxastro0} \\
    \hline
   	F1			& $0.46 \pm 0.13$  & $0.61 \pm 0.09$ \\
	Track Score	& $0.41 \pm 0.11$ &	$0.61 \pm 0.09$ \\
	Change Score	& $ 0.06  \pm 0.06$ & $0.20 \pm 0.09$ \\
	SCOT		& $0.17 \pm0.11$ &	$0.41 \pm 0.11$ \\
   \hline
  \end{tabular}
  \label{tab:comp}
  \vspace{-3pt}
\end{table}

\subsection{Feature Correlations}\label{sec:corr}

\begin{figure}
    \centering
    \begin{tabular}{cc}
    \includegraphics[width=0.38\linewidth]{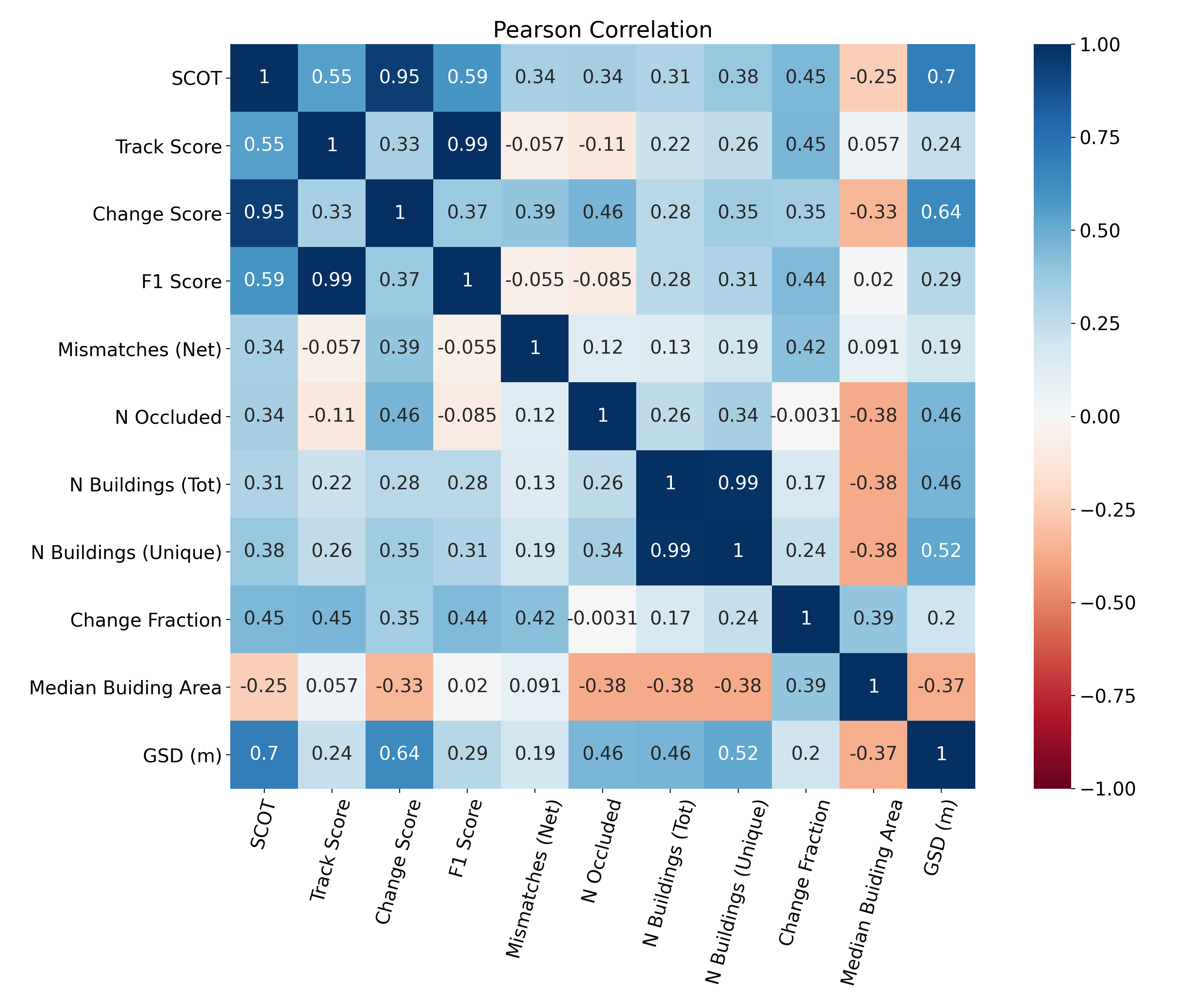} &   
    \includegraphics[width=0.49\linewidth]{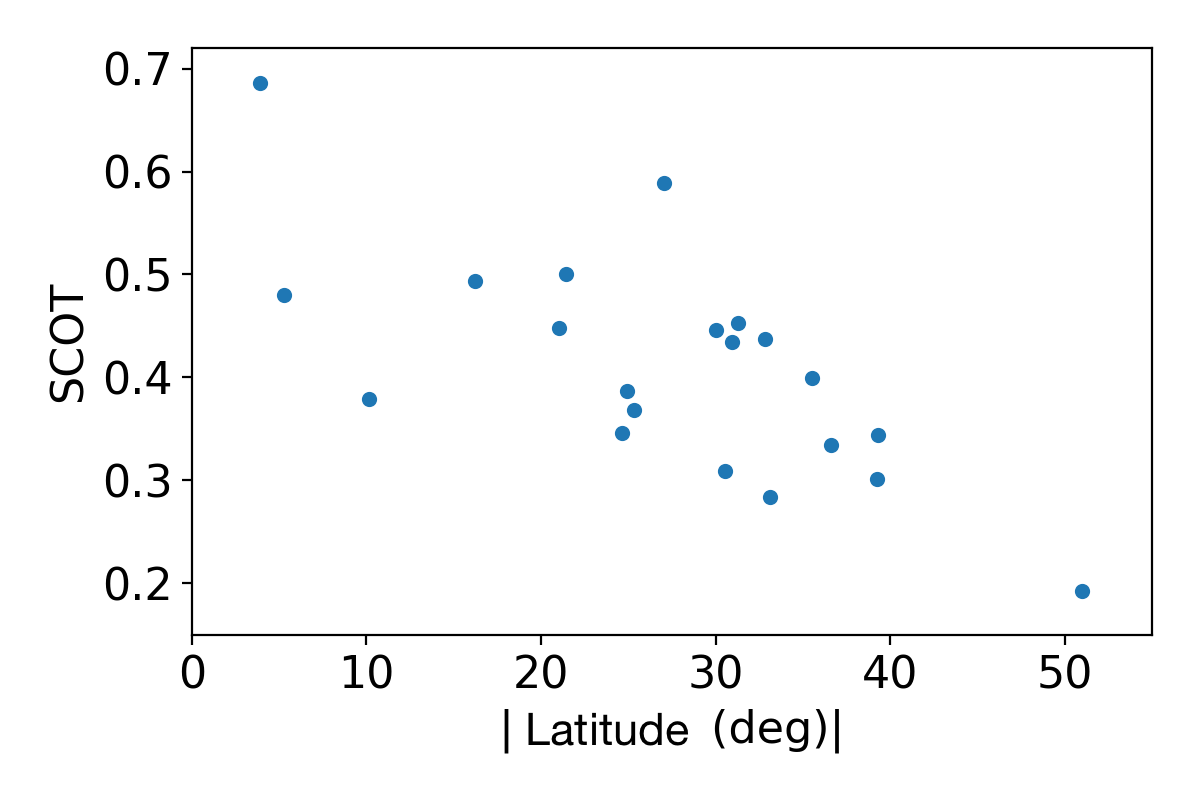} \\ 
        (a) Pearson Correlations & (b) SCOT vs latitude \\ [-0pt] 
    \end{tabular}
    \vspace{-5pt}
    \caption{ Correlations (a) and scatter plot (b) for the winning submission.}
    \label{fig:corr}
    \vspace{-1pt}
\end{figure}

Multiple features of the dataset and winning prediction that are worth exploring.  Figure \ref{fig:corr}a displays the correlation between various variables across the AOIs for the winning submission.  Most variables are positively correlated with the total SCOT score.  Note the high correlation between SCOT and the change score; 
since change detection is much more difficult this term ends up dominating. 

There are a number of intriguing correlations in Figure \ref{fig:corr}a, but one unexpected finding was the high (+0.7) correlation between ground sample distance (GSD), and SCOT.  This correlation is even stronger than the correlation between SCOT and F1 or SCOT and track score.  GSD is the pixel size of the imagery, so a higher GSD corresponds to larger pixels and lower resolution.  Furthermore, since all images are the same size in pixels (1024 $\times$ 1024), a larger GSD will cover more physical area, thereby increasing the density of buildings.  Therefore, one would naively expect an inverse correlation between GSD and SCOT where increasing GSD leads to decreased SCOT, instead of the positive correlation of Figure \ref{fig:corr}a.  

As it turns out, the processing of the SpaceNet 7 Planet imagery results in GSD $\approx$ 4.8m $\times$ Cos(Latitude).  Therefore latitude (or more precisely, the absolute value of latitude) is negatively correlated  with tracking (-0.39), change (-0.65) and SCOT (-0.70) score.  Building footprint tracking is apparently more difficult at higher latitudes, see Figure \ref{fig:corr}b.

The high negative correlation (-0.65) between the change detection term (change score) and latitude is noteworthy.  
Evidently, identifying building change is significantly harder at higher latitudes.  We leave conclusive proof of the reason for this phenomenon to further studies, but hypothesize that the reason is due to the greater seasonality and more shadows/worse illumination (due to more oblique sun angles) at higher latitudes.  
Figure \ref{fig:bloob}b
illustrates some of these effects.  Note the greater shadows and seasonal change than in Figure \ref{fig:bloob}a.  For reference, the change score for Figure \ref{fig:bloob}a (latitude of 20 degrees) is 0.30, whereas the change score for Figure \ref{fig:bloob}b (latitude of 40 degrees) is  0.09.

\subsection{Performance Curves}

\begin{figure}[b]
    \centering
    \includegraphics[width=0.46\linewidth]{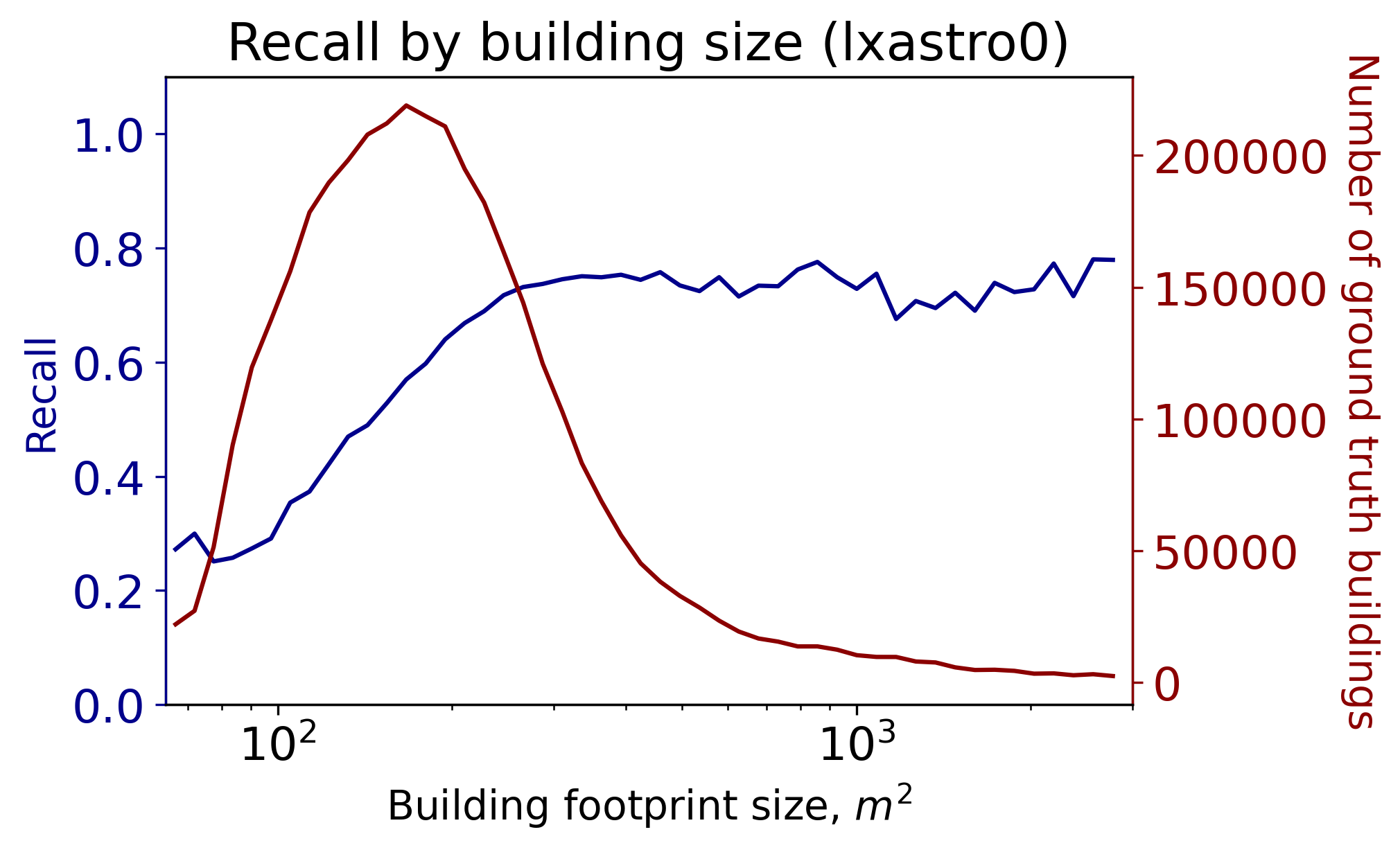}
    \vspace{-5pt}
    \caption{ Building recall as a function of area for the winning submission (IoU $\geq$ 0.25). }
    \label{fig:area}
    \vspace{-5pt}
\end{figure}

Object size is an important predictor of detection performance, as noted in a number of previous investigations (e.g.  \cite{yolt}). We follow the lead of analyses first performed in SpaceNet 4 \cite{sn4_area}  
  (and later SpaceNet 6 \cite{sn6_area})
  in exploring object detection performance as function of building area.  Figure \ref{fig:area} shows performance for all 4.4 million building footprints in the SpaceNet 7 public and private test sets for the winning submission of team lxastro0.

\begin{figure}[t]
    \centering
    \includegraphics[width=0.9\linewidth]{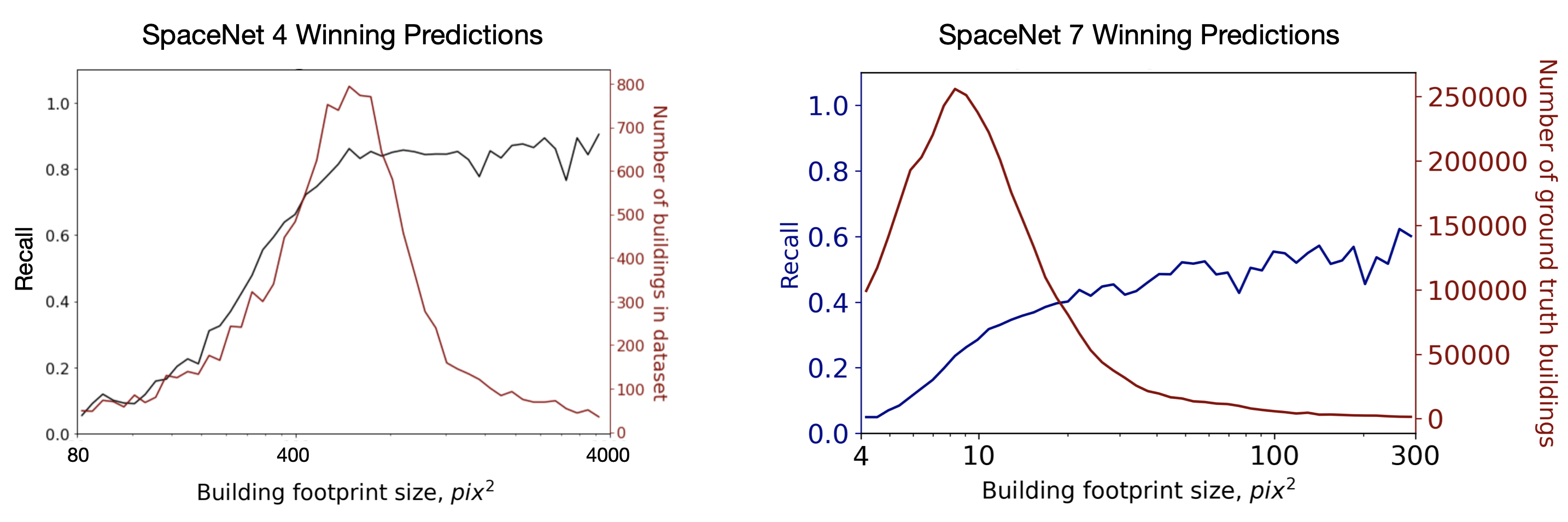}
    \vspace{-5pt}
    \caption{ Prediction performance as a function of building pixel area (IoU $\geq$ 0.5). }
    \label{fig:area2}
    \vspace{-1pt}
\end{figure}

The pixel size of objects is also of interest, particularly in comparison to previous SpaceNet challenges.  The SpaceNet 4 Challenge used 0.5m imagery, so individual pixels are 1/64 the area of our 4m resolution SpaceNet 7 data, yet for SpaceNets 4 and 7 the physical building sizes are similar \cite{sn7_area}. 
Figure \ref{fig:area2} plots pixel sizes directly (for this figure we adopt IoU $\geq$ 0.5 for direct comparisons), demonstrating the far superior pixel-wise performance of SpaceNet 7 predictions in the small-area regime ($\sim5\times$ greater for 100 pix$^2$ objects), though SpaceNet 4 predictions have a far higher score ceiling.   The high SpaceNet 7 label fidelity (see Figure \ref{fig:omni}) may help explain the over-achievement of the winning model prediction on small buildings.  SpaceNet 7 labels encode extra information not obvious to humans in the imagery, which models are apparently able to leverage.  Of course there is a limit (hence the score ceiling of SpaceNet 7 predictions), but this extra information does appear to help models achieve surprisingly good performance on difficult, crowded scenes.

\subsection{SCOT Analysis}

\begin{figure}
    \centering
    \begin{tabular}{cc}
    \includegraphics[width=0.4\linewidth]{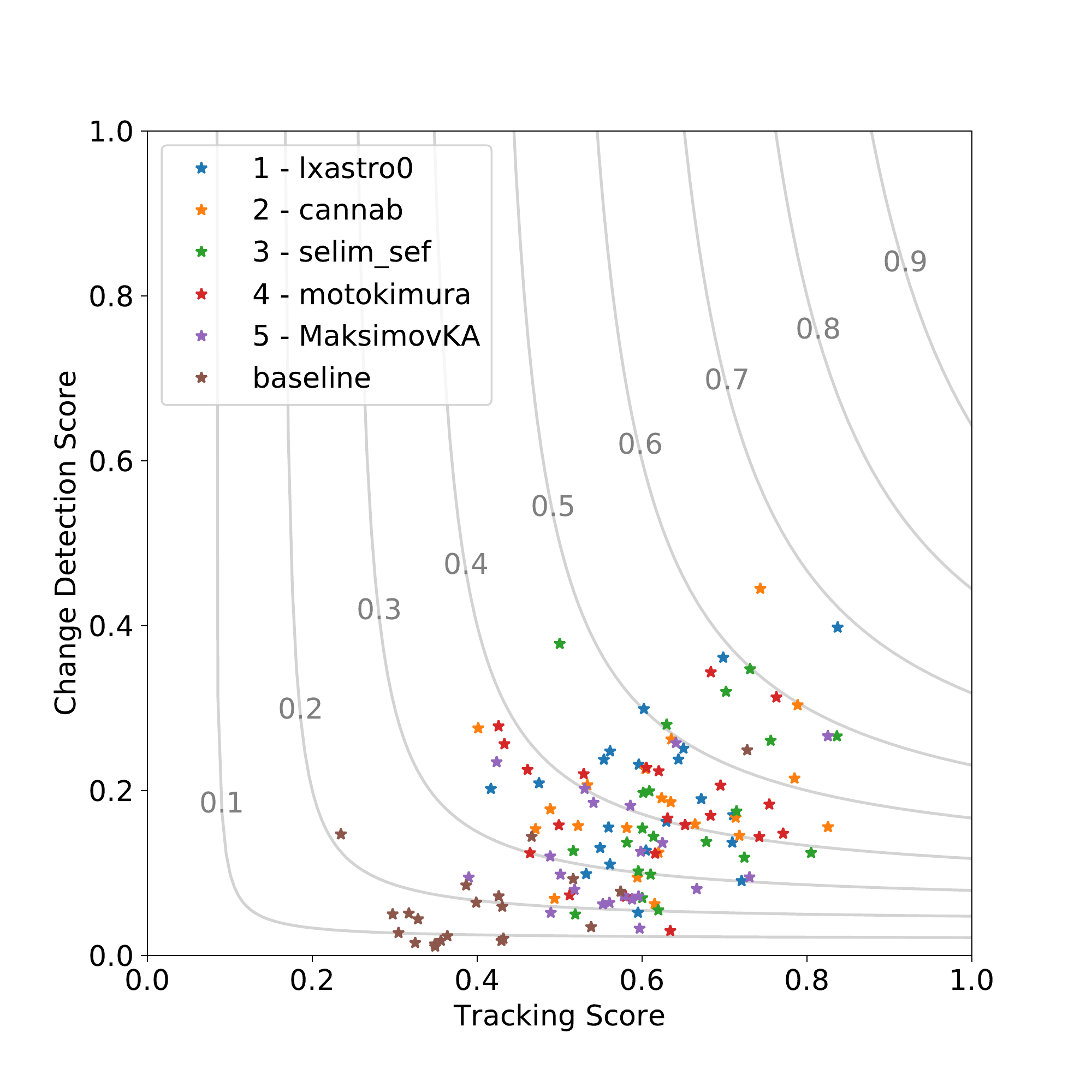} &   
    \includegraphics[width=0.4\linewidth]{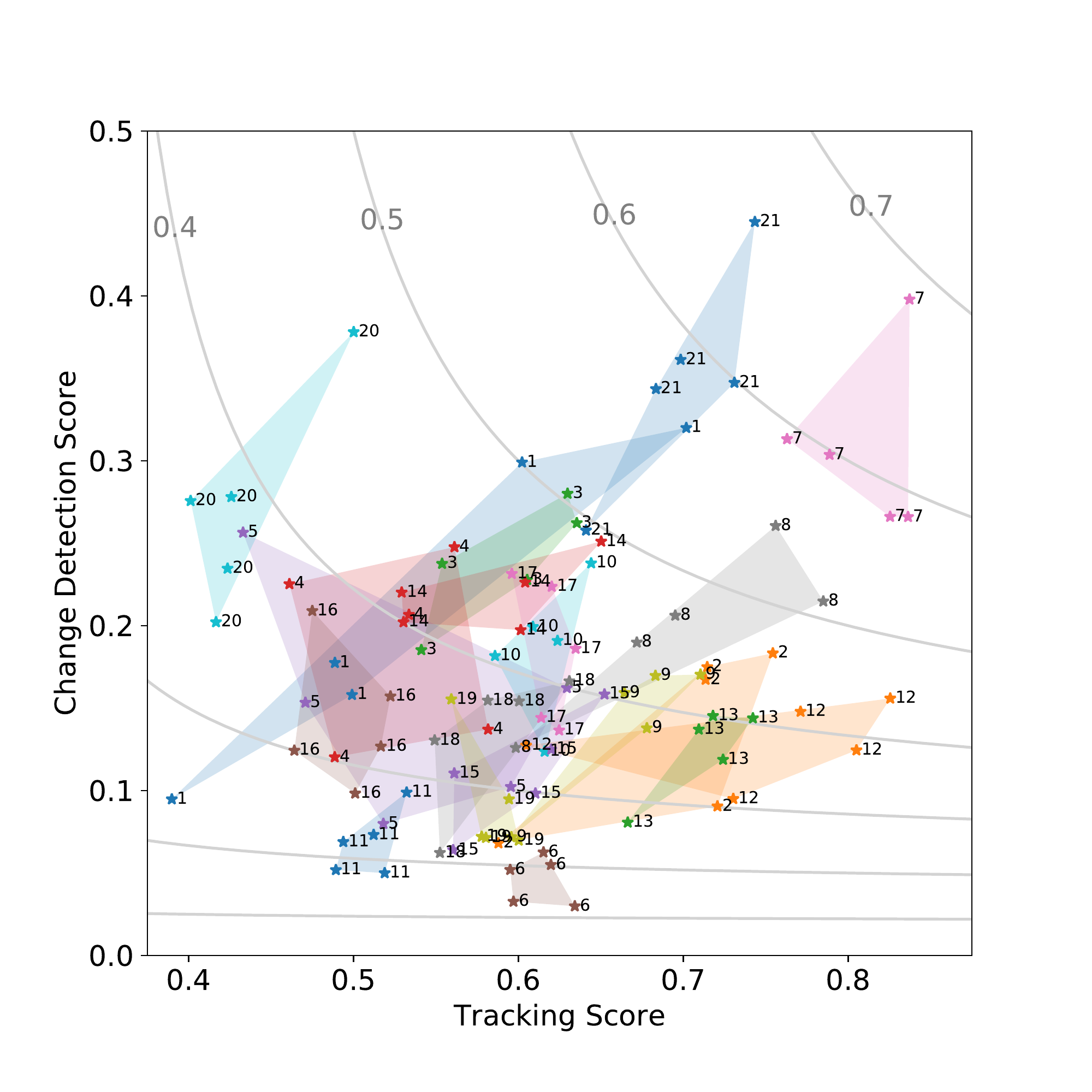} \\ 
    (a) By model & (b) By AOI \\ [-0pt] 
    \end{tabular}
    \vspace{-5pt}
    \caption{ Change score vs. tracking score for each combination of model and AOI, color-coded (a) by model and (b) by AOI.  Contour lines indicate SCOT score.}
    \label{fig:scotscores}
    \vspace{-1pt}
\end{figure}

\begin{figure}
    \centering
    \begin{tabular}{cccc}
    \includegraphics[width=0.2\linewidth]{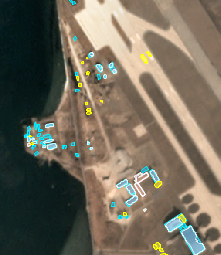} &   
    \includegraphics[width=0.2\linewidth]{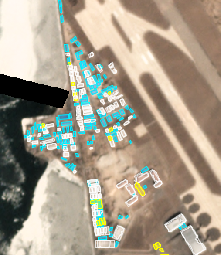} &
    \includegraphics[width=0.2\linewidth]{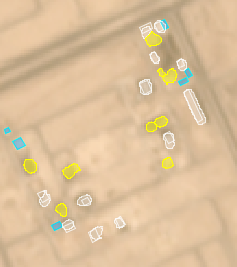} &   
    \includegraphics[width=0.2\linewidth]{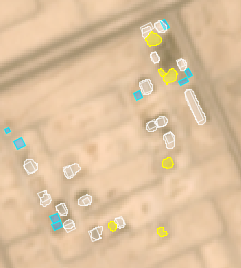} \\ 
    (a) AOI 20 before & (b) AOI 20 after & (c) AOI 12 before & (d) AOI 12 after \\ [-0pt] 
    \end{tabular}
    \vspace{-5pt}
    \caption{Detail of AOI 20 (a) before and (b) after the completion of new construction, and similarly for AOI 12 (c) before and (d) after.  Matched footprints are in white, false positives in yellow, and false negatives in blue.}
    \label{fig:scotexamples}
    \vspace{-5pt}
\end{figure}

Comparing the performance of the various models can give insight into the role played by the two terms that make up the SCOT metric.  Figure \ref{fig:scotscores}a plots change detection score against tracking score for each model in Table \ref{tab:results}, showing a weak correlation.  Breaking down those points by AOI in Figure \ref{fig:scotscores}b shows that deviations from linearity are largely model-independent, instead relating to differences among AOIs.  The AOIs labeled ``20'' and ``12'' show extreme cases of this variation (Figure \ref{fig:scotexamples}).  AOI 20 achieves a high change detection score despite a low tracking score because many buildings are detected either from first construction or not at all.  AOI 12, on the other hand, achieves a high tracking score despite a low change detection score because predicted building footprints often appear earlier than ground truth, potentially an effect of construction activity.  Such cases show the value in using both terms to make SCOT a holistic measure of model performance.

\section{Conclusions}

The winners of The SpaceNet 7 Multi-Temporal Urban Development Challenge all managed impressive performance given the difficulties of tracking small buildings in medium resolution imagery.  The winning team submitted by far the most and rapid (and therefore the most useful) proposal.  By executing a ``temporal collapse'' and identifying temporal step functions in footprint probability, the winning team was able to vastly improve both object tracking and change detection performance.  Inspection of correlations between variables unearthed an unexpected decrease in performance with increasing resolution.  Digging into this observation unearthed that the latent variable appears to be latitude, such that SCOT performance degrades at higher latitudes. We hypothesize that the greater lighting differences and seasonal foliage change of higher latitudes complicates change detection. 
Predictions for the SpaceNet 7 4m resolution dataset perform surprisingly well for very small buildings.  In fact, Figure \ref{fig:area2} showed that prediction performance for 100 pix$^2$ objects is $\sim5\times$  for SpaceNet 7 than for SpaceNet 4.  The high fidelity ``omniscient'' labels of SpaceNet 7 seem to aid models for very small objects, though the lower resolution of SpaceNet 7 results in a lower performance ceiling for larger objects.  Insights such as these have the potential to help optimize collection and labeling strategies for various tasks and performance requirements.  

Ultimately, the open source and permissively licensed data and models stemming from SpaceNet 7 have the potential to aid efforts to improve mapping and aid tasks such as 
emergency preparedness assessment, disaster impact prediction, disaster response, high-resolution population estimation, and myriad other urbanization-related applications.


\newpage

\bibliographystyle{splncs}
\bibliography{bib}

\end{document}